\documentclass[runningheads]{llncs}

\usepackage{iciap}

\usepackage{iciapabbrv}

\usepackage{graphicx}
\usepackage{booktabs}
\usepackage{url}
\usepackage{subcaption}
\usepackage{wrapfig}
\usepackage{comment}
\usepackage{dutchcal}
\usepackage{array}
\usepackage{float}
\usepackage[inline]{enumitem}
\usepackage{amsmath}
\usepackage{makecell}
\usepackage{comment}
\usepackage{wrapfig}
\usepackage{authblk}
\usepackage{bbding}  

\usepackage[accsupp]{axessibility}  

\usepackage{hyperref}

\usepackage{orcidlink}

\begin{document}

\title{7Bench: a Comprehensive Benchmark for Layout-guided Text-to-image Models} 

\titlerunning{7Bench}

\author{Elena Izzo\thanks{These authors contributed equally to this work and are listed in alphabetical order.}\inst{1}\Envelope\orcidlink{0000-0003-1105-4162} \and
Luca Parolari$^\star$\inst{1}\orcidlink{0000-0001-8574-4997} \and
Davide Vezzaro$^\star$\inst{1,2}\orcidlink{0009-0007-3139-6342} \and 
Lamberto Ballan\inst{1}\orcidlink{0000-0003-0819-851X}}

\authorrunning{E.~Izzo$^\star$, L.~Parolari$^\star$, D.~Vezzaro$^\star$, and L.~Ballan}

\institute{{Department of Mathematics, University of Padova, Padova, Italy} \and
Brain Technologies Innovation Division, Brain Technologies srl, Torino, Italy 
\email{\{elena.izzo, luca.parolari, davide.vezzaro.4\}@studenti.unipd.it} \email{lamberto.ballan@unipd.it}
}

\maketitle

\begin{abstract}
Layout-guided text-to-image models offer greater control over the generation process by explicitly conditioning image synthesis on the spatial arrangement of elements.
As a result, their adoption has increased in many computer vision applications, ranging from content creation to synthetic data generation.
A critical challenge is achieving precise alignment between the image, textual prompt, and layout, ensuring semantic fidelity and spatial accuracy. 
Although recent benchmarks assess text alignment, layout alignment remains overlooked, and no existing benchmark jointly evaluates both. 
This gap limits the ability to evaluate a model's spatial fidelity, which is crucial when using layout-guided generation for synthetic data, as errors can introduce noise and degrade data quality.
In this work, we introduce 7Bench, the first benchmark to assess both semantic and spatial alignment in layout-guided text-to-image generation.
It features text-and-layout pairs spanning seven challenging scenarios, investigating object generation, color fidelity, attribute recognition, inter-object relationships, and spatial control.
We propose an evaluation protocol that builds on existing frameworks by incorporating the layout alignment score to assess spatial accuracy.
Using 7Bench, we evaluate several state-of-the-art diffusion models, uncovering their respective strengths and limitations across diverse alignment tasks. The benchmark is available at \hyperlink{https://github.com/Elizzo/7Bench}{https://github.com/Elizzo/7Bench}.
\end{abstract}

\section{Introduction}

Layout-guided text-to-image models extend the standard text-to-image generation paradigm by introducing explicit layout control.
Among the various layout representations, bounding boxes have emerged as the most commonly adopted format due to their simplicity and ease of annotation.
This format enables users to intuitively specify the position, scale, and appearance of objects, offering a more controllable generation process.
Such fine-grained control over the image content fosters new opportunities for leveraging synthetic data, consequently increasing the demand and usage of layout-guided diffusion models in other computer vision tasks, \eg synthetic dataset generation~\cite{DBLP:conf/icpr/ParolariIB24, DBLP:conf/eccv/KupynR24}.

Despite the high potential of these tools, recent studies have identified several limitations~\cite{DBLP:conf/wacv/GrimalBFT24}.
In fact, models often struggle to generate images that accurately adhere to the given layout and textual prompt, especially when given uncommon spatial arrangements~\cite{DBLP:conf/iccv/XieLHLZ0S23}.
Additionally, they inherit fundamental issues from diffusion models, including catastrophic neglect, attribute binding, and attribute leaking, which affect both text and layout alignment.

Quantifying and identifying errors in text-image and layout-image faithfulness is crucial to properly evaluate and compare different layout-guided diffusion models. 
However, benchmarks collected for evaluating textual \textit{and} layout alignment are still missing.
As a result, the strategies adopted by the researchers to evaluate layout-guided generative models are heterogeneous and mainly based on traditional text-to-image benchmarks~\cite{DBLP:conf/iccv/Bakr0SKLE23,DBLP:conf/iccv/HuLKWOKS23, DBLP:conf/nips/SahariaCSLWDGLA22}. 
These include tests for text-image alignment in scenarios involving colors, attributes, or relations, but either lack any dedicated evaluation of layout alignment or suffer from inconsistencies among different evaluation protocols. 
For instance, layout alignment is assessed using a random subset of annotated datasets for object detection or grounding~\cite{DBLP:conf/cvpr/LiLWMYGLL23, DBLP:conf/wacv/ChenLV24, DBLP:conf/cvpr/PhungGH24} or by researchers collecting their own sets of box-and-text pairs~\cite{DBLP:conf/iccv/XieLHLZ0S23}. 
A recent effort~\cite{DBLP:conf/cvpr/0001LYGWB22} introduces the evaluation of spatial skills (\eg position, size or shape) but overlooks semantic evaluation. 
This lack of a unified approach leads to a significant misalignment among studies in terms of evaluation metrics, the number of samples investigated, and the specific layout skills being tested.

In this paper, we propose 7Bench, a comprehensive benchmark for layout-guided models characterized by pairs of textual prompts and sets of bounding boxes to evaluate models on $7$ different scenarios. The aim is to investigate models' abilities in objects' generation, colors, attributes, relationships, and spatial control. We design a structured prompt collection annotated with bounding boxes, collecting a dataset with $224$ samples equally distributed in the $7$ scenarios. We introduce an evaluation method based on two complementary measures: the text-alignment and the layout-alignment scores. Our evaluation protocol extends an existing framework that focuses on text-image alignment by introducing an assessment of layout accuracy.
We systematically evaluate state-of-the-art layout-guided text-to-image diffusion models using our benchmark, comparing both fine-tuned models on grounding information and training-free approaches. Results obtained on 7Bench demonstrate room for improvement in all models, validating the scientific rationale of our approach.

In summary, our main contributions include: 
\begin{enumerate*}[label=(\roman*)]
    \item 7Bench, a novel benchmark for layout-guided text-to-image models, designed to evaluate both text-image and layout-image alignment across seven diverse and challenging scenarios;
    \item a meticulous collection of 224 text and bounding boxes pairs, ensuring a broad coverage of various semantic and spatial arrangements;
    \item an evaluation protocol based on two complementary measures: text-alignment and layout-alignment scores, which assess the alignment of generated images with respect to both textual descriptions and spatial layouts; and
    \item a comprehensive evaluation of four state-of-the-art layout-guided text-to-image diffusion models using 7Bench, providing insights into the strengths and weaknesses of different approaches, including a comparison between fine-tuned and training-free models.
\end{enumerate*}

\section{Related Works}

\textbf{Benchmarks for Text-to-image Generation.} Text-to-image generation is the process of creating images from textual descriptions. The rise of diffusion models~\cite{DBLP:conf/nips/HoJA20} revolutionised this task, outperforming traditional GAN-based approaches~\cite{DBLP:conf/icml/ReedAYLSL16} in generating high-quality and diverse images.
Despite the outstanding performance, these models often struggle to produce images that accurately align with the description,  failing to bind attributes or specific details to objects~\cite{DBLP:conf/wacv/GrimalBFT24}.
As a result, benchmarks and evaluation metrics have been introduced to measure the faithfulness of a generated image to its text input on various challenging scenarios: colors, attributes, counting, relations, or objects~\cite{DBLP:conf/iccv/HuLKWOKS23, DBLP:conf/nips/HuangSXLL23, DBLP:conf/iccv/Bakr0SKLE23, DBLP:conf/nips/WuYHRA24, DBLP:conf/iclr/FengHFJANBWW23, DBLP:conf/nips/SahariaCSLWDGLA22}. 
However, with the introduction of layout-guided text-to-image models, all these benchmarks miss two key elements: the input layouts of the object described in the textual prompt and an evaluation metric to quantify the layout alignment.

\noindent
\textbf{Layout-guided Text-to-image Diffusion Models.} 
Layout-guided text-to-image diffusion models incorporate predefined spatial layouts for finer control over image generation.
Generally, these methods inject grounding information into pre-trained models by adding new trainable layers or tokens~\cite{DBLP:conf/cvpr/LiLWMYGLL23, DBLP:conf/cvpr/YangWGLLWD0L0W23}, or manipulating cross-attention layers at inference time to steer the generation process without requiring additional training~\cite{DBLP:conf/iccv/XieLHLZ0S23, DBLP:conf/wacv/ChenLV24, DBLP:conf/cvpr/PhungGH24}. Evaluating these models requires assessing both text and layout alignment, but the absence of comprehensive benchmarks has led researchers to develop their own evaluation strategies.

\noindent
\textbf{Evaluation of Layout-guided Diffusion Models.}
The evaluation of layout-guided models is often a two-step framework assessing text and layout faithfulness separately. 
Text alignment is evaluated by benchmarks from the traditional text-to-image domain.
Instead, layout alignment is often assessed by generating images from bounding boxes and captions sampled from COCO2014~\cite{DBLP:conf/eccv/LinMBHPRDZ14} or Flickr30K Entities~\cite{DBLP:conf/iccv/PlummerWCCHL15} validation sets. The generation quality is then quantified with metrics like FID~\cite{DBLP:conf/nips/HeuselRUNH17} and YOLO score~\cite{DBLP:conf/iccv/LiWKTS21}.
However, this evaluation is heterogeneous among papers and does not follow a shared protocol. For instance, in~\cite{DBLP:conf/cvpr/LiLWMYGLL23} a subset of $30$k samples are generated while in~\cite{DBLP:conf/cvpr/PhungGH24} only $5$k.
Moreover, in~\cite{DBLP:conf/iccv/XieLHLZ0S23}, the authors point out the need for a distinct evaluation dataset, independent of COCO images, since it is commonly used to train the models.
In~\cite{DBLP:conf/cvpr/0001LYGWB22}, a benchmark for layout-guided image generation for spatial control skills is introduced.
However, it focuses only on spatial abilities without considering all the other challenging scenarios for generative models.
In this work, we introduce a comprehensive benchmark for layout-guided models, aiming to test them on various skills, from colors to spatial layouts.

\begin{figure}[t]
\centering
\includegraphics[width=\textwidth]{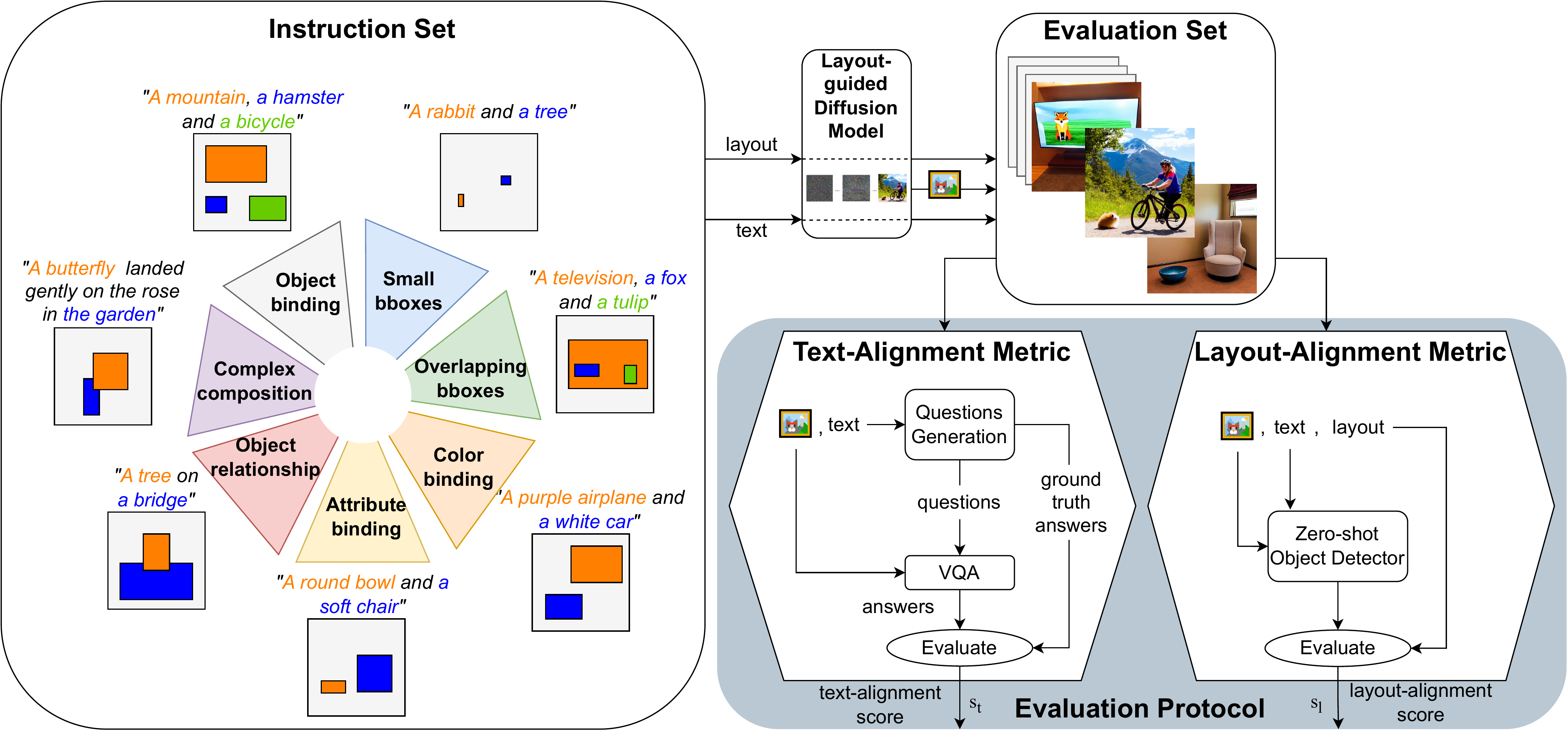}
\caption{7Bench overview. Our benchmark contains prompts from $7$ different scenarios based on semantic and spatial arrangement characteristics. The evaluation protocol to apply on the set of generated images is based on two metrics: text- and layout-alignment scores to evaluate text and layout faithfulness, respectively.}
\label{fig:benchmark}
\end{figure}

\section{7Bench Benchmark}

In this section, we present 7Bench, a benchmark specifically designed to evaluate the performance of layout-guided text-to-image diffusion models.
It assesses how faithfully the generated image adheres to both the input text and the specified layout by employing an evaluation protocol that measures alignment along both dimensions.
The benchmark comprises 224 diverse text prompts, each paired with a set of bounding boxes that define the intended spatial arrangement of objects in the image.
Samples are carefully curated to probe the capabilities of layout-aware generative models under a range of challenging conditions.
Overall, the benchmark spans 7 distinct scenarios, categorized based on the complexity and structure of the textual descriptions and associated layouts.
An overview of the benchmark is illustrated in Fig.~\ref{fig:benchmark}.

\subsection{Prompt and Bounding-Box Collection}
\label{subsec:dataset}
We created a benchmark of $224$ samples. 
Each sample is a pair made by a textual prompt and a set of bounding boxes (one for each object in the prompt). We divided the $224$ samples into seven scenarios based on the characteristics of their textual prompts and bounding boxes: object binding, small bboxes, overlapping bboxes, color binding, attribute binding, object relationship, and complex composition. We collected 32 prompts for each scenario to carefully balance the distribution and do not under/over-represent one specific case.

The sets of textual prompts are generated following a template whose formalism is inspired by the disentangle representation theory~\cite{DBLP:conf/iccv/TragerPZABS23}. Each prompt contains \textit{N} objects, each of which can be qualified by an attribute or be in relation to another object. Hence, the object at position \textit{i} in a prompt is a token belonging to the set $\mathcal{O}$, it can be qualified by an attribute $\mathcal{a}_i$ in the set $\mathcal{A}$ or in relation $\mathcal{r}_{ij}$ to another object at position \textit{j}, where the relation is in the set $\mathcal{R}$.

\begin{figure}[t]
    \centering
    \includegraphics[width=0.85\textwidth]{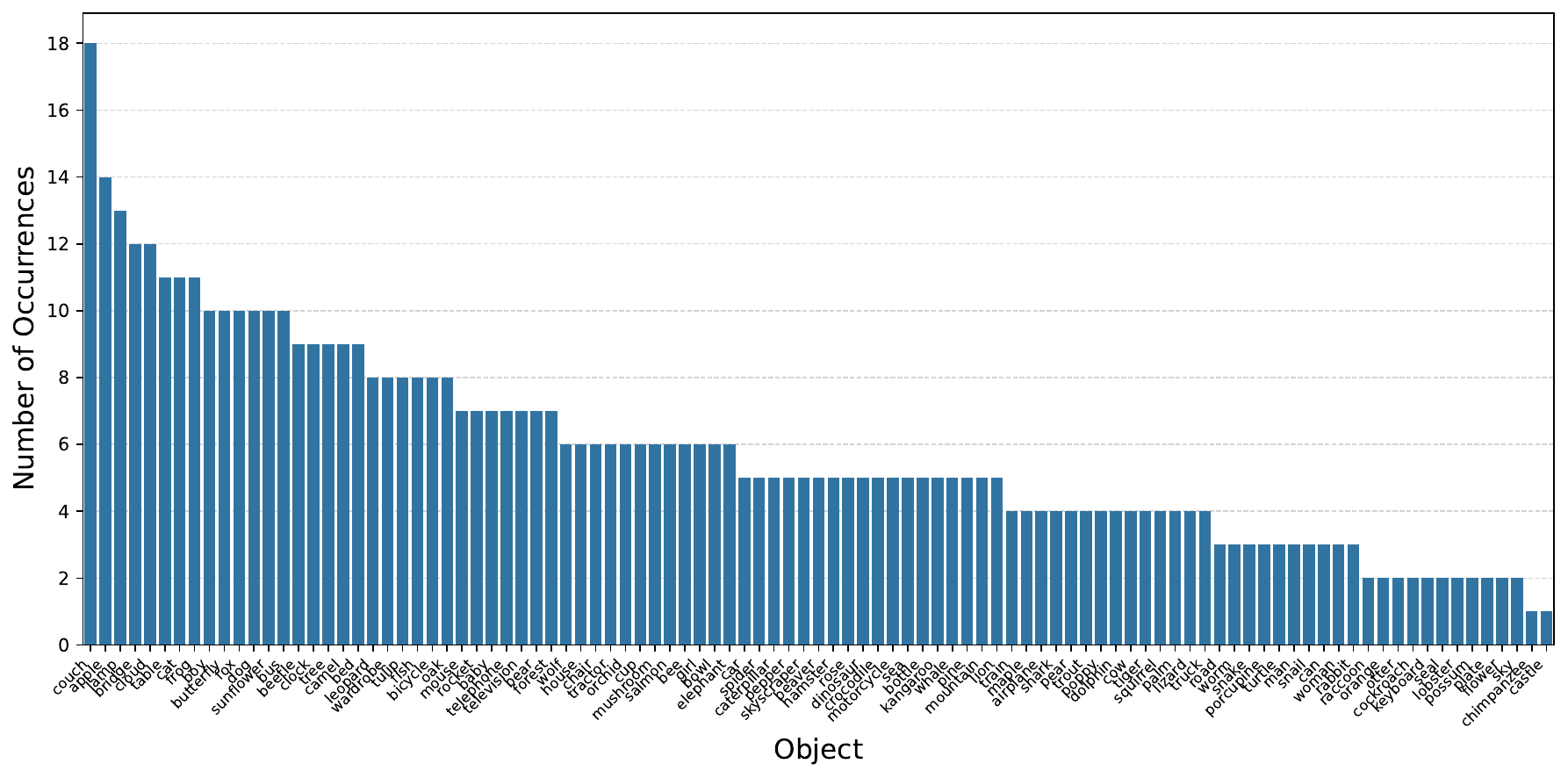}
    \caption{Number of occurrences by object computed across all prompts in 7Bench.}
    \label{fig:objects-distribution}
\end{figure}

For example, for a prompt with \textit{N}$=2$ objects, the template \textit{t} is:
\begin{equation}
    \textit{t} = \text{"\textit{det}($\mathcal{o}_1$,$\mathcal{a}_1$) $\mathcal{a}_1$ $\mathcal{o}_1$ \textit{rel}(\textit{and}, $\mathcal{r}_{12}$) \textit{det}($\mathcal{o}_2$,$\mathcal{a}_2$) $\mathcal{a}_2$ $\mathcal{o}_2$"}
\end{equation}
where, $\mathcal{o}_i \in \mathcal{O}$, $\mathcal{a}_i \in \mathcal{A}$, \textit{det}($\mathcal{o}_i$,$\mathcal{a}_i$) is a determinant that depends on the object or attribute if present, and \textit{rel}(\textit{and}, $\mathcal{r}_{ij}$) is the coordinating conjunction \textit{and} or a relation $\mathcal{r}_{ij} \in \mathcal{R}$ if present.
We vary the number of objects \textit{N} in \{$1,2,3,4$\}. Fig.~\ref{fig:objects-distribution} shows the objects' distribution across all collected prompts.

In order to define the layout requirements, the position of each object $\mathcal{o}_i$ in a textual prompt is defined by a bounding box $\textit{bbox}_i$. In particular, the desired position of the object $\mathcal{o}_i$ in the generated image is defined by a set of coordinates $\textit{bbox}_i = [x_{min}^i,y_{min}^i,x_{max}^i,y_{max}^i]$, where $0 \leq x_{min} < x_{max} \leq 1$ and $0 \leq y_{min} < y_{max} \leq 1$.
Unless otherwise specified, each bounding box is manually collected by arranging a realistic layout for each prompt that maintains proportions between objects and does not contain overlaps.

Hereafter, we illustrate each scenario investigated in our benchmark, describing the characteristics of their textual prompts and bounding boxes.

\subsubsection{Object Binding.}
This scenario aims to study the catastrophic neglect in the layout-guided models and the right localization of each generated object. Catastrophic neglect is a phenomenon that characterizes the text-conditioned diffusion models and consists in not generating or mixing together the elements described in the prompt. For this scenario, we collect a set of samples varying the number of objects \textit{N} in $\{1,2,3,4 \}$. 
We do not add any attribute or relation. The resulting textual prompt is constructed following the template:
\begin{equation}
    t_{obj} = \text{"\textit{det}($\mathcal{o}_1$) $\mathcal{o}_1$, \textit{det}($\mathcal{o}_2$) $\mathcal{o}_2$, \textit{det}($\mathcal{o}_3$) $\mathcal{o}_3$  and \textit{det}($\mathcal{o}_4$) $\mathcal{o}_4$"}.
\end{equation}

\begin{wrapfigure}{r}{7cm}
    \centering
    \includegraphics[width=\linewidth]{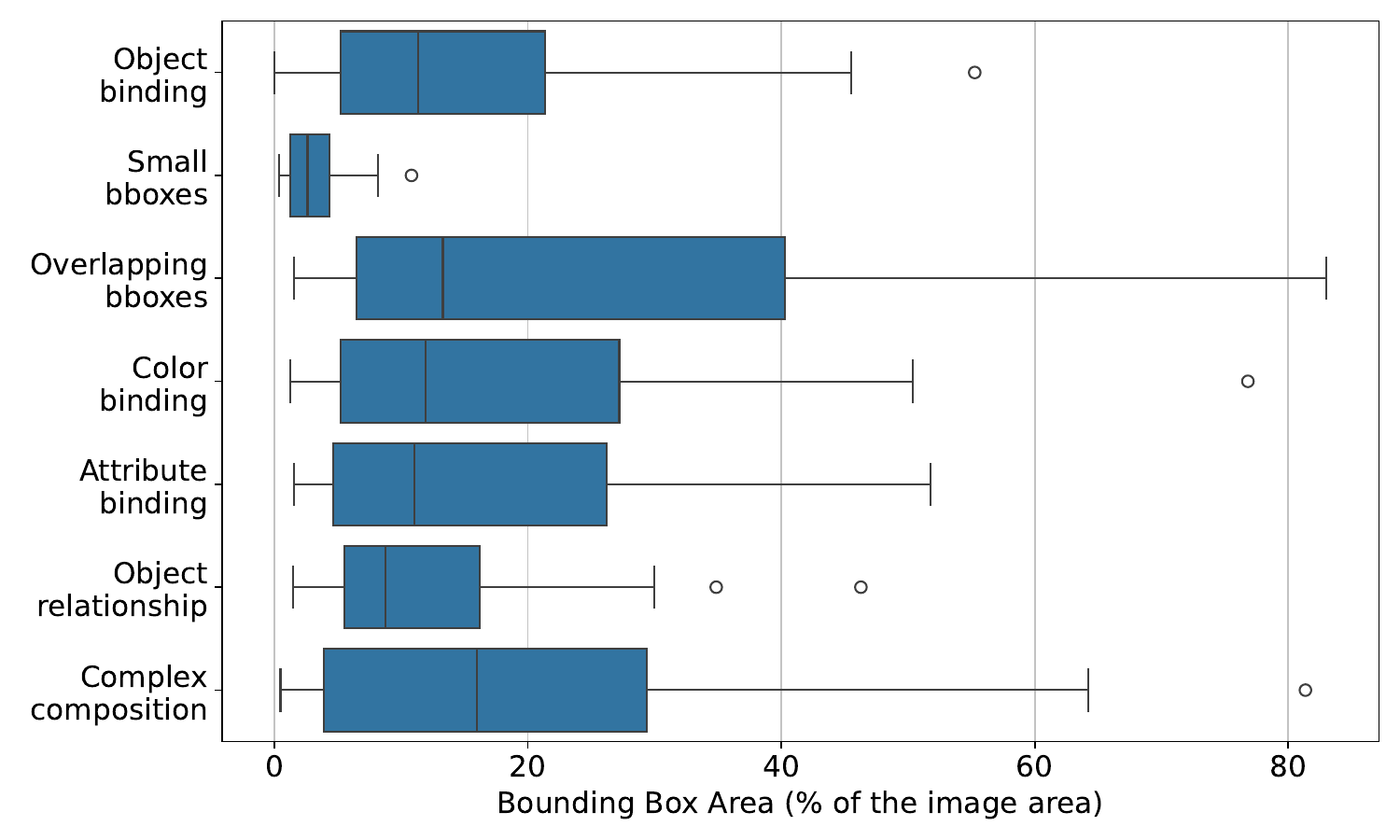}
    \caption{Distribution of bounding boxes area per scenario. Values are displayed percentage with respect to the image size.}
    \label{fig:bboxes-distribution}
\end{wrapfigure} 
\subsubsection{Small Bboxes.}
The use of small bounding boxes as the input represents a critical challenge for layout-guided text-to-image diffusion models.
Due to their limited spatial extent, small bounding boxes often fail to provide sufficient conditioning signals during the denoising process, leading to inaccurate object placement or complete omission~\cite{patel2024enhancing}.
In order to investigate generative abilities under such conditions, we introduce the \textit{small bboxes} scenario.
It is characterized by a spatial layout where the bounding boxes's area is between $3$\% and $10$\% of the total image area. 
Mathematically, given a bounding box $\textit{bbox}_i = [x_{min}^i,y_{min}^i,x_{max}^i,y_{max}^i]$, its area $A_i$ is defined as $A_i = (x_{max}^i - x_{min}^i)\cdot(y_{max}^i-y_{min}^i)$. 
Given an $w \times h$ image and its area $A_{image} = w \cdot h$, then $A_i$ is randomly chosen following the constraint $0.03 \cdot A_{image} \leq A_{image} \cdot A_i \leq 0.1 \cdot A_{image}$.
Fig.~\ref{fig:bboxes-distribution} shows the distribution of the areas of the bounding box collected in this scenario in comparison with the other ones. The textual template is $t_{obj}$ from the \textit{object binding} scenario.

\subsubsection{Overlapping Bboxes.} 
Similar to the \textit{small bboxes} scenario, another critical and relevant aspect to investigate in a layout-guided diffusion model is the use of overlapping bounding boxes as guidance. Those can be used to specify background and foreground layers or to specify details on an object. As previously done, we follow the textual template $t_{obj}$, that represent 1-4 objects without attributes or relations. All the bounding boxes in a sample overlap at least one other bounding box in the same sample. Specifically, $\forall i \in \{1,...,N\} \quad \exists j_{\in \{1,...,N \}}\neq i | A_i \cap A_j > 0$.

\subsubsection{Color Binding.}
This scenario aims to investigate the adherence of color-based guidance on depicted objects. 
Despite color being the most commonly used attribute for describing objects in scenes, state-of-the-art text-conditioned diffusion models still demonstrate issues in applying the correct color to the target object. 
For studying this characteristic in layout-guided text-to-image diffusion models, we define a set $\mathcal{C}$ of eleven universal basic colors based on the findings of Berlin and Kay~\cite{berlin1991basic}: \textit{black}, \textit{blue}, \textit{brown}, \textit{gray}, \textit{green}, \textit{pink}, \textit{purple}, \textit{red}, \textit{white}, \textit{yellow}, \textit{orange}.
The textual template report a color attribute for each object described in the prompt: 
\begin{equation}
t_{color} = \text{"\textit{det}($\mathcal{c}_1$) $\mathcal{c}_1$ $\mathcal{o}_1$, \textit{det}($\mathcal{c}_2$) $\mathcal{c}_2$ $\mathcal{o}_2$, \textit{det}($\mathcal{c}_3$) $\mathcal{c}_3$ $\mathcal{o}_3$  and \textit{det}($\mathcal{c}_4$) $\mathcal{c}_4$ $\mathcal{o}_4$"}.
\end{equation}

\subsubsection{Attribute Binding.} 
This scenario investigates the capacity of a generative model to correctly bind an object and its attribute.
It extends the \textit{color binding} scenario by adding several attributes related to various types: color, shape, material, appearance, and dimension of the object. In total, we investigate $30$ different attributes setting $\mathcal{A}$ = \{  \textit{aggressive, black, blue, bright, clean, crowded, dark, fast, fluffy, fuzzy, green, happy, large, pink, red, rotten, rough, shiny, short, silver, small, smooth, snowy, soft, tall, warm, white, wooden, yellow}\}. For this scenario, we avoid the use of relations and we vary the number of objects \textit{N} in $\{1,2,3,4 \}$:
\begin{equation}
t_{attr} = \text{"\textit{det}($\mathcal{a}_1$) $\mathcal{a}_1$ $\mathcal{o}_1$, \textit{det}($\mathcal{a}_2$) $\mathcal{a}_2$ $\mathcal{o}_2$, \textit{det}($\mathcal{a}_3$) $\mathcal{a}_3$ $\mathcal{o}_3$  and \textit{det}($\mathcal{a}_4$) $\mathcal{a}_4$ $\mathcal{o}_4$"}.
\end{equation}

\subsubsection{Object Relationship.} 
The relationships between objects are crucial when describing complex scenes, but is also a difficult task for generative models.
For this reason, we introduce the \textit{object relationship} scenario. 
It is characterized by a textual prompt with $2$ or $4$ objects correlated to each other by a relation, and does not use attributes.
Inspired by studies on spatial cognition~\cite{landau1993whence}, we define a set $\mathcal{R}$ of 11 relations: $\mathcal{R} = $ \{\textit{above, below, beside, far from, near, next to, on, over, to the left of, to the right of, under} \}. The textual template is defined as:
\begin{equation}
t_{rel} = \text{"\textit{det}($\mathcal{o}_1$) $\mathcal{r}_{12}$ \textit{det}($\mathcal{o}_2$) $\mathcal{o}_2$ and \textit{det}($\mathcal{o}_3$) $\mathcal{r}_{34}$ \textit{det}($\mathcal{o}_4$) $\mathcal{o}_4$"}.
\end{equation}

\subsubsection{Complex Composition.} 
In order to test layout-guided models with more open-world and challenging compositional textual prompts and layout, we collect $32$ samples in which to combine all the previous investigated scenarios. Hence, in the \textit{complex composition} scenario, the textual prompts can contain from $1$ to $4$ objects. Each object is qualified by at least one attribute that describes its color, shape, material, appearance, or dimension, and it is in relation to another object. The bounding boxes associated with each object can have an area smaller than $10$\% of the generated image and can overlap each other.

\subsection{Evaluation Protocol}
\label{subsec:eval_protocol}

To evaluate the alignment between the generated images and the input prompt, we introduce two complementary measures: the text-alignment score $s_{text}$ and the layout-alignment score $s_{layout}$. These scores quantify the image adherence to both the semantic and spatial constraints specified in the prompt.

\subsubsection{Text Faithfulness.}
We define the text-alignment score $s_{text} \in [0, 1]$ as the TIFA score~\cite{DBLP:conf/iccv/HuLKWOKS23}, a widely used metric for evaluating semantic alignment in text-to-image models. 
This score represents the proportion of correct answers produced by a Vision Question Answering (VQA) model when analyzing the generated images. 
The evaluation process involves generating a set of questions and corresponding answers based on the input prompt. 
These are created by a Large Language Model (LLM), ensuring that the assessment remains independent of the image generation process. 
The VQA model is then queried with these questions while processing the generated image, and its responses are compared against the expected answers from the LLM. 
The final score $s_{text}$ is computed as the accuracy of the VQA model’s answers with respect to the reference answers provided by the LLM. 
A higher $s_{text}$ value indicates strong semantic consistency between the image and the prompt.

\subsubsection{Layout Faithfulness.}
We define the layout-alignment score $s_{layout} \in [0, 1]$ as the Area Under Curve (AUC) of accuracy@k values computed over a range of Intersection over Union (IoU) thresholds. The $s_{layout}$ score captures the spatial accuracy of objects' placement within the generated images.

Specifically, given a generated image $\hat{I}$, the set of objects $\{o_i\}_{i=1}^{N}$ and the set of target bounding boxes representing the ground truth position of objects $\{t_i\}_{i=1}^{N}$, we apply an object detector (\eg OWL-ViT~\cite{DBLP:journals/corr/abs-2205-06230}) to obtain a set of $M$ detections associated with their corresponding label and confidence score: $D = \{(d_j, l_j, c_j)\}_{j=1}^M$.
For each target object $o_i$, we filter the set of detections by matching the label with the object, thus $\hat{D}_i = \{(d_j, l_j, c_j) \in D \mid l_j = o_i\}$.
Among the filtered detections $\hat{D}_i$, we select the bounding box with the higher score $\hat{d}_i = d_{j^*}$ with $j^* = \arg\max_{j:\, (d_j, l_j, c_j) \in \hat{D}_i} c_j$.
We then compute the IoU between the selected detection $\hat{d}_i$ and the target bounding box $t_i$, denoted as $\text{IoU}_i = \text{IoU}(t_i, \hat{d}_i)$.
Subsequently, these IoU values are thresholded at multiple levels $k \in \{0.1, 0.2, \ldots, 0.9\}$ to calculate \begin{equation}
accuracy@k = \frac{1}{N} \sum_{i=1}^N \mathbf{1}[\text{IoU}_i \geq k].
\end{equation} 
Then, the final layout-alignment score $s_{layout}$ is obtained as the area under the resulting accuracy@k curve.

\section{Experiments}

\subsection{Experimental Setup}

In this section, we present a systematic evaluation of 4 popular layout-guided text-to-image diffusion models on our 7Bench benchmark. The models are all open source and have been accurately chosen to explore a wide range of methods and techniques. In particular, we test GLIGEN (G)~\cite{DBLP:conf/cvpr/LiLWMYGLL23}, that is trained with grounding information, and three training-free approaches:  Attention Refocusing (G\_AR)~\cite{DBLP:conf/cvpr/PhungGH24}, BoxDiff (G\_BD)~\cite{DBLP:conf/iccv/XieLHLZ0S23}, and Cross Attention Guidance (SD\_CAG)~\cite{DBLP:conf/wacv/ChenLV24}. The first two are built on top of GLIGEN, the last one uses Stable Diffusion as the underlying model. Finally, we include Stable Diffusion v1.4 (SD)~\cite{DBLP:conf/cvpr/RombachBLEO22} in the analysis for a comparison in terms of textual alignment.

Given a model, we generate 16 images for each sample in 7Bench, varying the seed from $1$ to $16$. We get a total of $17.920$ synthetic images with dimension $512 \times 512$, equally distributed in the five models under investigation.

We evaluate the generated images using the evaluation protocol described in Section~\ref{subsec:eval_protocol}. We use pre-trained weights for TIFA and OWL-ViT~\footnote{\href{https://github.com/Yushi-Hu/tifa}{https://github.com/Yushi-Hu/tifa}}$^,$\footnote{\href{https://huggingface.co/docs/transformers/model\_doc/owlvit}{https://huggingface.co/docs/transformers/model\_doc/owlvit}}.

\subsection{Quantitative and Qualitative Evaluation}

\begin{figure}[t]
    \centering
    \begin{subfigure}[b]{0.49\textwidth}
        \centering
        \includegraphics[width=\textwidth]{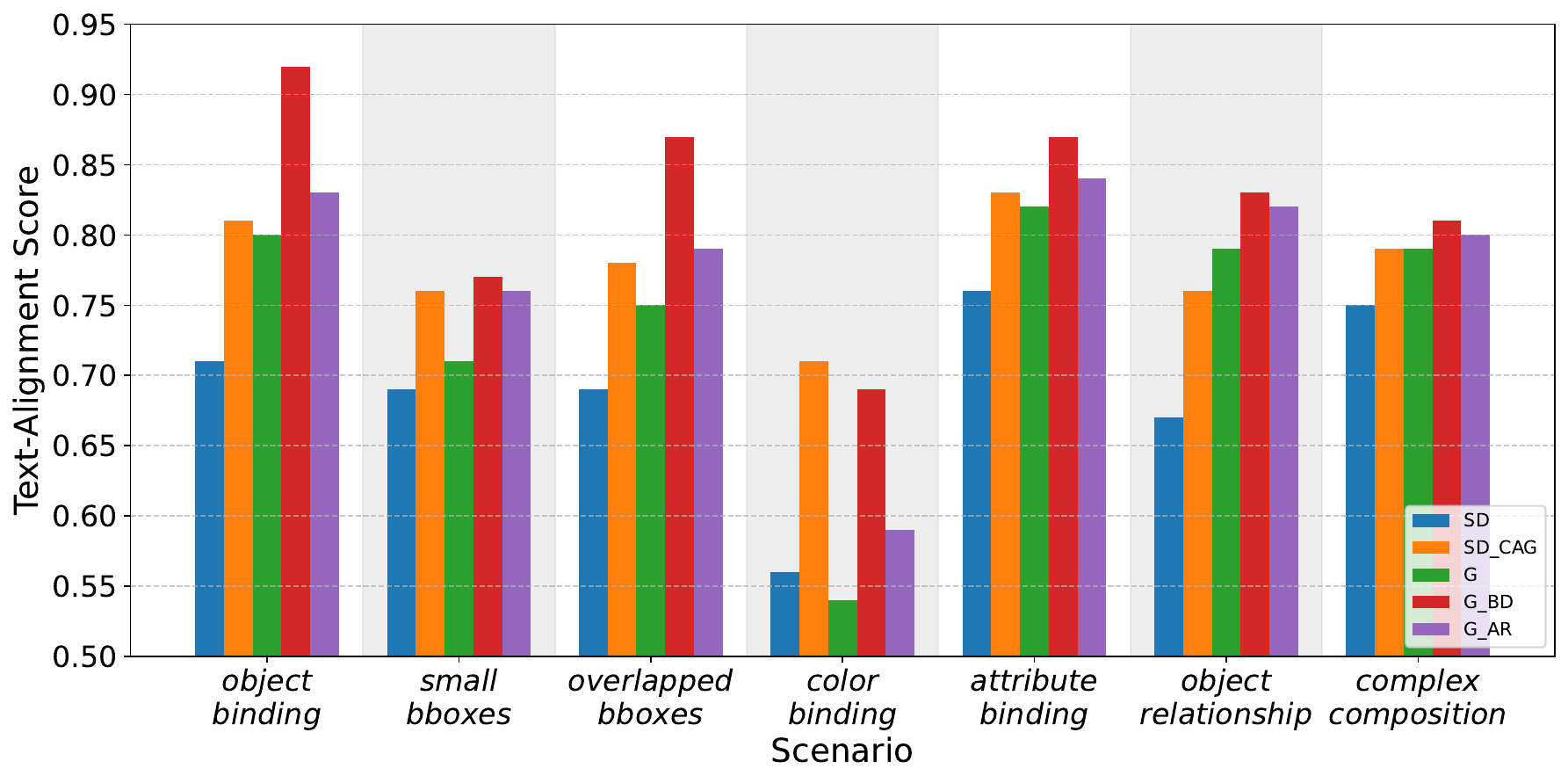}
        \caption{ }
        \label{fig:results-text}
    \end{subfigure}
    \hfill
    \begin{subfigure}[b]{0.49\textwidth}
        \centering
     \includegraphics[width=\textwidth]{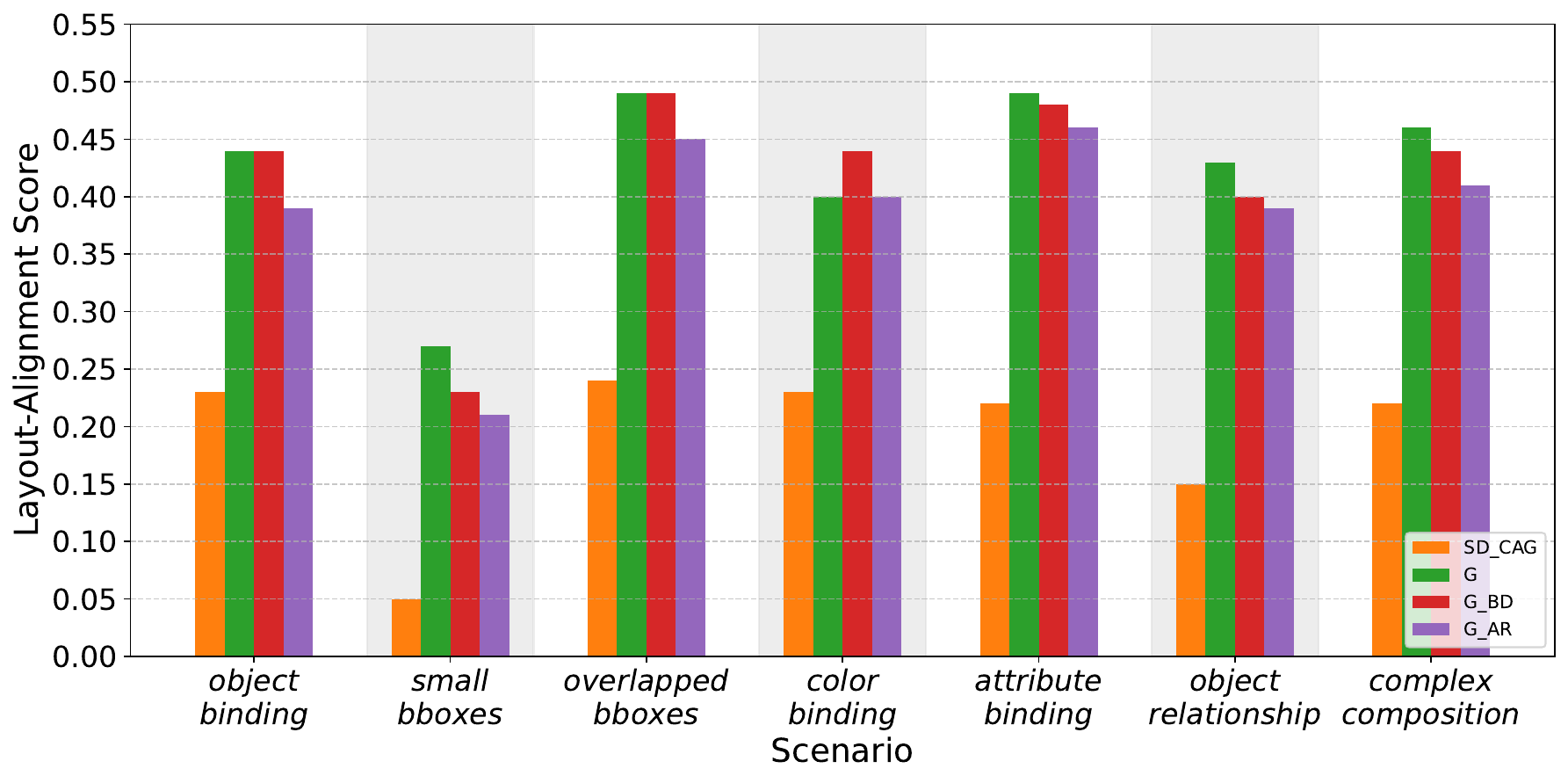}
        \caption{ }
        \label{fig:results-layout}
    \end{subfigure}
    \caption{Performance per scenario. Fig.~\ref{fig:results-text} and Fig.~\ref{fig:results-layout} show textual and layout alignment, respectively.}
    \label{fig:results}
\end{figure}

We first present an evaluation average over scenarios (Fig.~\ref{fig:results}), reporting both text-alignment (Fig.~\ref{fig:results-text}) and layout-alignment (Fig.~\ref{fig:results-layout}) results. We recall that the two metrics are complementary measures that aim to quantify text and layout alignment, respectively. Both scores take values in $[0,1]$, where $1$ means perfect alignment and $0$ means complete misalignment.

Depending on the model and scenario, the text alignment varies approximately from $0.55$ to $0.9$, and the layout alignment from $0.05$ to $0.5$, showing room for improvement in the current state-of-the-art models in all the scenarios in our benchmark. In terms of layout-alignment score, the curves show that methods significantly struggle to generate objects that respect a provided small bounding box. Instead, they seem to not struggle in \textit{overlapping bboxes}, often finding creative solutions to overlap objects as requested.
Overall, models based on a trained layout-guided model (GLIGEN, BoxDiff, and Attention Refocusing) obtain superior performance with respect to Cross Attention Guidance (based on SD). This is not surprising considering that GLIGEN gets the best results in the majority of the scenarios.

As concerns the text-alignment (Fig.~\ref{fig:results-text}), there is more scores' fluctuation both among models and scenarios. Overall, training-free approaches are good allies and can significantly boost the model to which they are added improving its concept knowledge. In particular, BoxDiff generally outperforms the other models, and SD\_CAG gets results comparable to those of the other training-free approaches despite being built on top of SD instead of GLIGEN. Interestingly, all layout-guided models converge to the same score in \textit{complex composition}, suggesting their comparable abilities when open-world and challenging compositional textual and layout are provided.

\begin{figure}[t]
    \centering
    \begin{subfigure}[b]{0.49\textwidth}
        \centering
        \includegraphics[width=\textwidth]{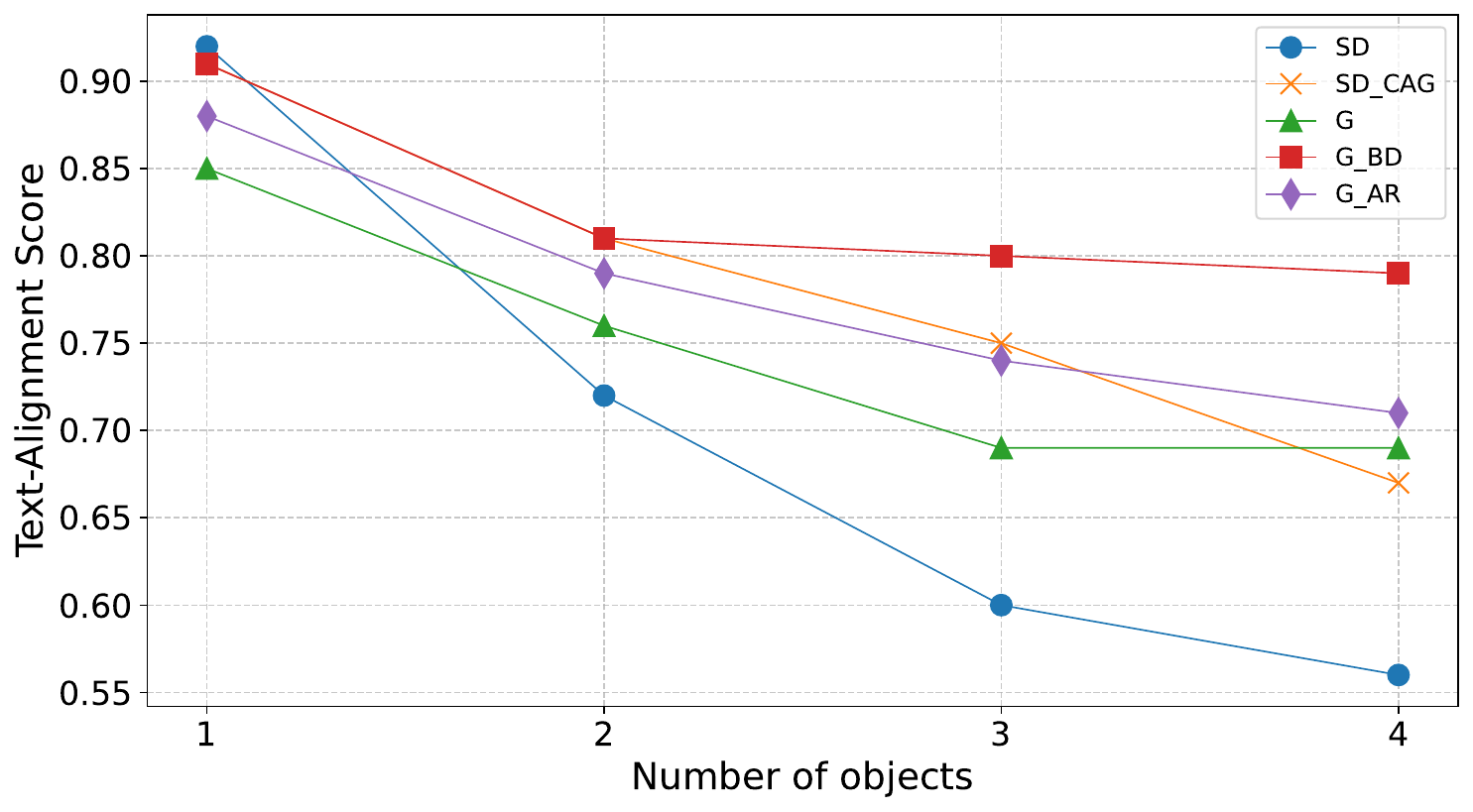}
        \caption{ }
        \label{fig:ablation-text}
    \end{subfigure}
    \hfill
    \begin{subfigure}[b]{0.49\textwidth}
        \centering
        \includegraphics[width=\textwidth]{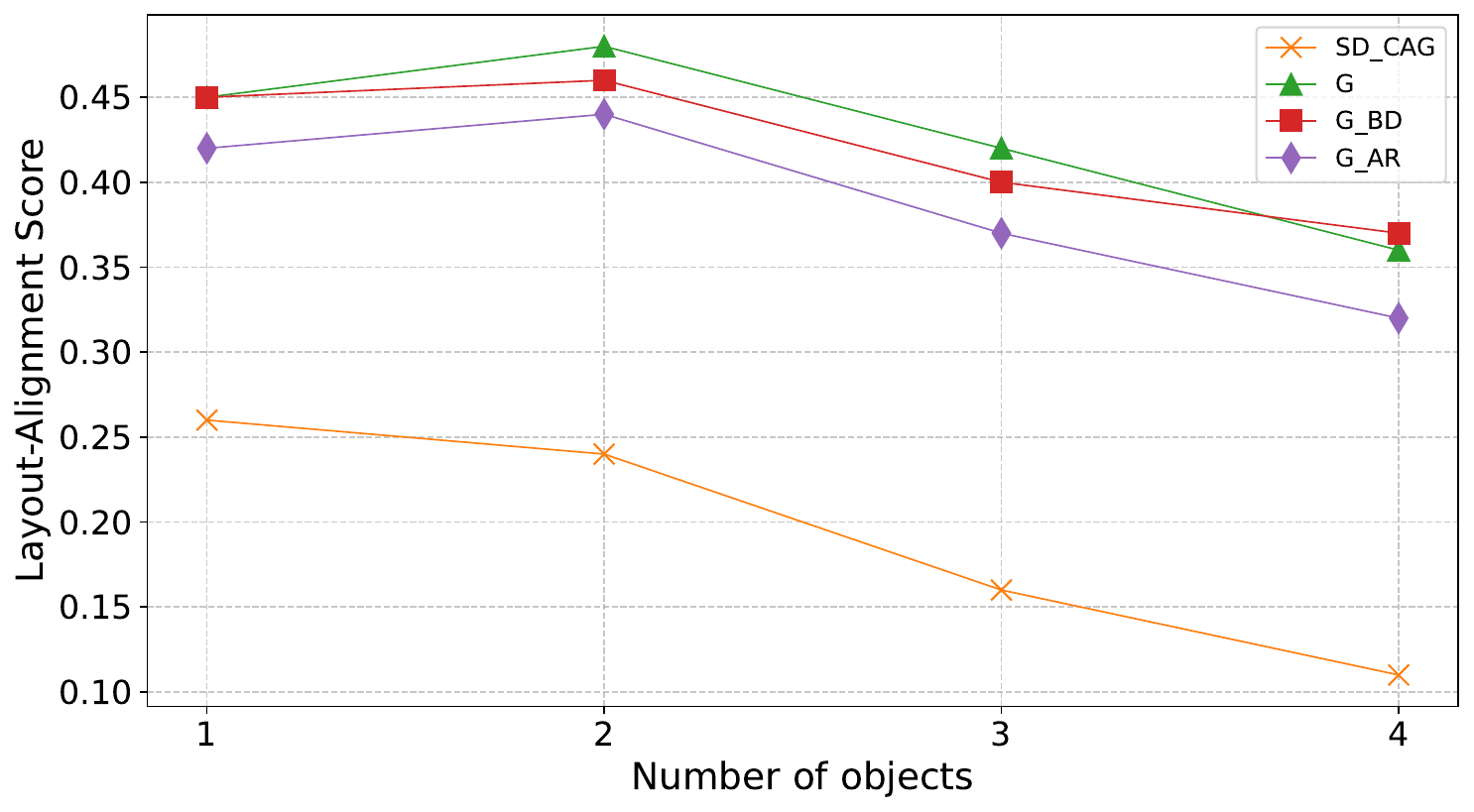}
        \caption{ }
        \label{fig:ablation-layout}
    \end{subfigure}
    \caption{Performance per number of objects in the prompt. Fig.~\ref{fig:ablation-text} and Fig.~\ref{fig:ablation-layout} show textual and layout alignment, respectively. }
    \label{fig:ablation}
\end{figure}

\begin{figure}[H]
    \centering
    \includegraphics[width=0.85\textwidth]{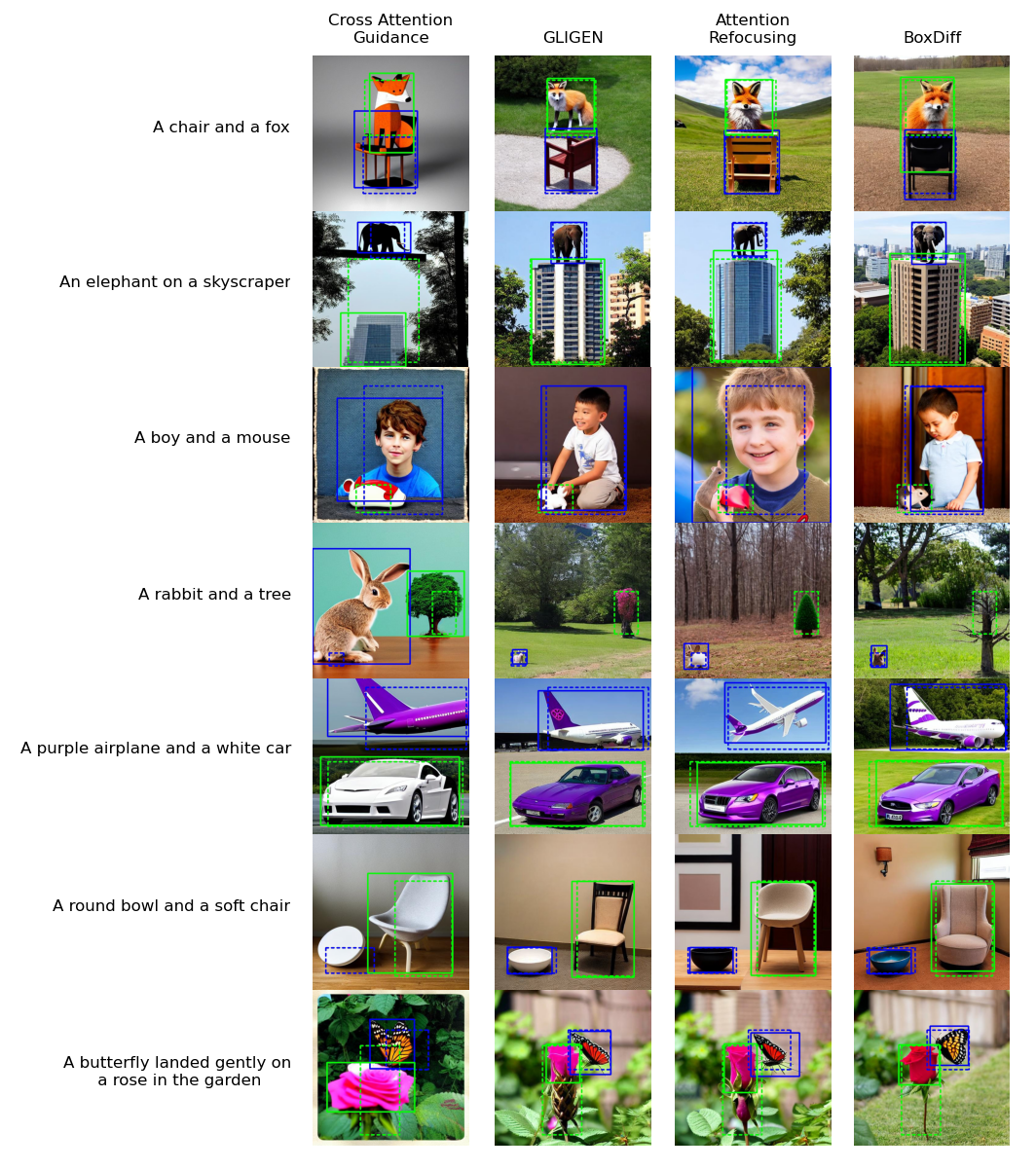}
    \caption{Qualitative results. Ground-truth object locations are indicated by dashed bounding boxes, while detected objects are outlined with solid bounding boxes. Colors identify different target objects.}
    \label{fig:qualitative}
\end{figure}

Secondly, we present the results averaged over the number of objects in the textual prompt (Fig.~\ref{fig:ablation}). Both text-alignment (Fig.~\ref{fig:ablation-text}) and layout-alignment (Fig.~\ref{fig:ablation-layout}) curves confirm the expected trend: a progressive degradation in performance increasing the number of objects in the prompt. In addition, models based on GLIGEN exhibit a lower degradation factor with respect to Cross Attention Guidance and Stable Diffusion.

Finally, Fig.~\ref{fig:qualitative} shows some qualitative results of all the layout-guided models under investigation on a subset of randomly selected prompts. In the images, we show the input bounding box of each object in the textual prompt and the filtered bounding box obtained by applying our Layout Faithfulness pipeline.

\section{Conclusion}

We introduced 7Bench, a benchmark tailored for evaluating layout-guided text-to-image diffusion models across diverse and challenging scenarios. Unlike prior evaluations that focus solely on text alignment or adopt inconsistent strategies for layout assessment, 7Bench provides a unified protocol to measure both text-image and layout-image faithfulness. Our results highlight the limitations of training-free approaches and the importance of fine-tuning for effective spatial control. We believe that 7Bench will support more rigorous comparisons, drive the development of more controllable generative models, and enable progress in downstream tasks where structured generation is essential.

\section*{Acknowledgments}

We acknowledge the CINECA award under the ISCRA initiative, for the availability of high performance computing resources and support.

%
%
\bibliographystyle{splncs04}
\bibliography{main}
\end{document}